\documentclass[letterpaper, 10 pt, conference]{ieeeconf}  

\IEEEoverridecommandlockouts                              

\renewcommand\vec{\mathbf}

\usepackage[ruled]{algorithm2e}

\usepackage{graphics} 
\usepackage{url}
\usepackage{booktabs}

\usepackage{amsmath} 
\usepackage{amsfonts}
\usepackage{amssymb}  
\usepackage{multicol}
\usepackage{multirow}
\usepackage{graphicx}
\usepackage{svg}
\usepackage{layouts}
\usepackage{wrapfig}
\usepackage{caption}
\usepackage{graphicx,subcaption}
\usepackage{easyReview}
\usepackage{printlen}
\usepackage{bm}
\usepackage{siunitx}
\usepackage{colortbl}	
\usepackage[utf8]{inputenc}
\usepackage{pgf}
\usepackage{tikz}
\usepackage{tikzscale}
\usepackage{pgfplots}
\DeclareUnicodeCharacter{2212}{\textendash}
\pgfplotsset{compat=1.14}
\usepgfplotslibrary{external}
\usepgfplotslibrary[external]
\tikzset{external/force remake}

\usepackage{makecell}
\usepackage{hyperref}

\usepackage{tcolorbox}
\newcommand{\namedblock}[2]{%
  \begin{tcolorbox}[
    title=#1,
    colback=white,
    colframe=black,
    boxrule=0.5pt,
    left=0pt,
    right=0pt,
    top=0pt,
    bottom=0pt,
    boxsep=3pt,
    width=0.85\linewidth,
    fonttitle=\bfseries,
    colbacktitle=white!50!white,
    coltitle=black,
  ]
    #2
  \end{tcolorbox}
}

\title{\LARGE \bf
Safe Learning of Locomotion Skills from MPC
}

\author{Xun Pua $^{1}$
 , Majid Khadiv$^{1}$ 
\thanks{$^{1}$Munich Institute of Robotics and Machine Intelligence, Technical
University of Munich, Germany {\tt\small xun.pua@tum.com, majid.khadiv@tum.de}}
}

\begin{document}

\maketitle
\thispagestyle{empty}
\pagestyle{empty}


\begin{abstract}
Safe learning of locomotion skills is still an open problem. Indeed, the intrinsically unstable nature of the open-loop dynamics of locomotion systems renders naive learning from scratch prone to catastrophic failures in the real world. In this work, we investigate the use of iterative algorithms to safely learn locomotion skills from model predictive control (MPC). In our framework, we use MPC as an expert and take inspiration from the safe data aggregation (SafeDAGGER) framework to minimize the number of failures during training of the policy. Through a comparison with other standard approaches such as behavior cloning and vanilla DAGGER, we show that not only our approach has a substantially fewer number of failures during training, but the resulting policy is also more robust to external disturbances.
\end{abstract}

\section{INTRODUCTION}
Ensuring safe, robust and reliable locomotion control for quadrupedal robots in dynamic environments is a significant challenge. In multi-contact scenarios, two dominant approaches for controlling legged robots are optimal control via model predictive control (MPC) and (deep) reinforcement learning (DRL). While recent works demonstrated that MPC can produce robust behaviors for multi-gait quadrupedal locomotion \cite{Meduri.19012022}\cite{mastalli2023agile}\cite{grandia2023perceptive}, their run-time computation is intensive and handling uncertainties in particular at contact events is highly challenging \cite{tassa2011stochastic}\cite{drnach2021robust}\cite{hammoud2021impedance}. 
DRL, on the other hand, avoids run-time computation issues as the policy training happens in an offline phase, and can handle any type of uncertainties through domain randomization. However, DRL is highly sample inefficient, the underlying algorithms for continuous control are potentially unstable, and intensive reward shaping for each individual task is required \cite{ha2024learning}. Furthermore, current DRL algorithms do not take safety issues into account, which makes it difficult to use them on the real hardware to fine-tune the policies trained in simulation.

One promising approach to combining the benefits of optimal control and learning is through the use of MPC to guide the search for learning optimal control policies \cite{carius2020mpc}\cite{viereck2022valuenetqp}\cite{Bogdanovic.7142021}\cite{Khadiv.2023}\cite{jenelten2024dtc}. The work of \cite{Bogdanovic.7142021}\cite{jenelten2024dtc} use trajectory optimization to guide the exploration of the DRL. While this can simplify the reward shaping problem, it is still sample-inefficient and no safety mechanism is in place for potential learning directly on the hardware. 

To avoid problems with DRL in generating optimal policies, \cite{carius2020mpc}\cite{viereck2022valuenetqp}\cite{Khadiv.2023} use MPC as an expert to learn a control policy. In particular, \cite{carius2020mpc} uses the data from MPC to learn the control Hamiltonian function, which is then used at run-time to generate safe actions. While this approach handles the constraints that can be interpreted as safety, still massive computation at run-time is required to produce actions from the learned control Hamiltonian function. To reduce the run-time computation, \cite{viereck2022valuenetqp} proposes to learn the value function of the MPC that is then used to solve a one-step optimal control problem. At run time, a simple quadratic program is solved, for which highly efficient off-the-shelf solvers exist. To further reduce the computation time, \cite{Khadiv.2023} showed that by using a proper representation of the action space it is possible to learn directly a policy that mimics the actions of the MPC, from a dataset generated offline (i.e. behavioral cloning, BC). They also showed that they can generate a wide range of gaits on a quadruped and also can learn robust policies directly in the sensor space. 

It is widely known that BC suffers from compounding errors, which can lead the robot to states out of the distribution of the training dataset, causing failure \cite{Ross.2010}. To address these limitations, we propose the utilization of online learning frameworks for legged locomotion control that accounts for the distribution shift during policy execution. Our proposed framework aims to enhance behavioral cloning by: 1) increasing safety during training which can facilitate learning or fine-tuning policies directly in the real world, 2) collecting more relevant data that can cover a more relevant part of the state space and hence improving sample efficiency, and 3) mitigating the effects of distribution shift which leads to more robust policies.

The main contributions of the paper are as follows:
\begin{itemize}
    \item We propose a new algorithm called LocoSafeDAGGER to safely learn locomotion policies from an expert controller. 
    \item Through an extensive set of simulation experiments, we compare our approach with BC and LocoDAGGER (an adapted version of DAGGER for locomotion) and show that our approach reduces substantially the number of failures and increases robustness against external disturbances, without sacrificing performance  
\end{itemize}

The remainder of the paper is structured as follows: In Section II, we give a brief overview of the existing algorithms that constitute the foundations of our work. Section III outlines our proposed framework. Section IV presents the implementation details. Section V presents the results and discussions. Finally, Section VI concludes our findings and outlines the future perspectives for the current work.

\section{Background}
\subsection{Behavioral cloning (BC)}
BC, a form of imitation learning, trains an agent to perform tasks by observing and mimicking a teacher's actions \cite{pomerleau1988alvinn}. In a typical BC workflow, a dataset $D$ is collected from an expert, comprising observed states $s$ and expert actions $a$. A policy $\pi$ is then trained using this data. The training objective can be formally described as $\pi = \arg\max_{\pi} \mathbb{E}_D[\log \pi(a \mid s)]$, which aims to maximize the expected log-likelihood of the policy's output actions given the input states observed in the dataset \cite{Hussein.2018}. This approach enables the policy to learn state-action associations that replicate the expert's demonstrated behavior. Due to its simplicity and ease of implementation, some recent studies have successfully trained and deployed policies for multi-gait legged locomotion control using BC \cite{Khadiv.2023}\cite{youm2023imitating}.

\subsection{Data Aggregation (DAGGER)}

While BC's simplicity and effectiveness in certain contexts make it valuable for legged robot control, it has the notable disadvantage of distribution shift. Ross and Bagnell \cite{Ross.2010} proved that if a BC-trained policy makes mistakes with probability $\epsilon$, the total mistakes over a rollout time of $T$ will be $T^2\epsilon$. This means that longer rollouts increase the likelihood of the policy encountering states not represented in the dataset, leading to potential failure. Simply collecting more expert data is not viable in legged robot control, as experts are less likely to make mistakes or encounter difficult-to-recover scenarios. This approach also has poor data efficiency and high computational cost. To address these issues, iterative supervised learning algorithms have been proposed. Unlike BC's offline training approach, iterative algorithms expand the dataset each iteration by collecting new data through policy rollouts.

One well-known iterative algorithm applying expert action relabelling is the DAGGER algorithm (refer to algorithm \ref{alg:dagger}) introduced by Ross et al. \cite{Ross.1122010}. The algorithm's core idea is that at each iteration, the policy $\hat{\pi}$ is rolled out to collect a trajectory of states $s$. The expert $\pi^*$ is then queried to obtain expert actions $a$ for these observed states. These new observations and expert actions are appended to the dataset $D$, and the policy is retrained to fit this expanded dataset. This approach actively enriches the dataset with situations the policy will actually encounter. To address policy instability during earlier iterations, a combination of the expert and policy is used, with the expert's influence decreasing as iterations increase. Ross et al. \cite{Ross.1122010} claim that this algorithm can outperform BC-trained policies in racing and Atari games.

\begin{algorithm}[htbp]
\caption{DAGGER}\label{alg:dagger}
\KwIn{Number of algorithm iterations $N$,\newline Expert influence $\beta$}
Initialize $D \gets \emptyset$\;
Initialize $\hat{\pi}_1$ to any policy in $\Pi$\;
\For{$i=1$ to $N$}{
	Let $\pi_i = \beta_i \pi^* + (1-\beta_i)\hat{\pi}_i$\;
	Sample T-step trajectories using $\pi_i$\;
	Get dataset $D_i = \{ (s,\pi^*(s)) \}$ of visited states by $\pi_i$ and actions given by expert\;
	Aggregate datasets: $D \gets D \cup D_i$\;
	Train policy $\hat{\pi}_{i+1}$ on $D$\;
 }
\Return{best $\hat{\pi}_i$ on validation}
\end{algorithm}

\subsection{SafeDAGGER}

Various versions of the DAGGER algorithm have been developed to address certain weaknesses of the original algorithm. One such variation is SafeDAGGER, introduced by Zhang and Cho \cite{Zhang.5202016} (refer to algorithm \ref{alg:safedagger}). At its core, SafeDAGGER follows the same process as vanilla DAGGER. The main difference lies in the policy rollouts. Instead of allowing the policy to continue in unsafe situations, the expert takes over. Subsequently, only states where the expert intervened are relabelled with expert actions and appended to the dataset. This approach ensures that only states not yet encountered by the policy are added to the training data, avoiding overloading the dataset with the same data over and over. Additionally, the algorithm prevents catastrophic failures during the policy rollout stage, making it more suitable for deployment on real robot hardware.

\begin{algorithm}[htbp]
\caption{SafeDAGGER}\label{alg:safedagger}
\KwIn{Number of algorithm iterations $N$,\newline Safety strategy $\pi_{safe}$}
Initialize $D$ with expert $\pi^*$\;
Pre-train $\pi_1$ on $D$\;
\For{$i=1$ to $N$}{
	Collect $D^\prime$ using $\pi_i$ and $\pi_{safe}$\;
	$D^\prime \gets \{ (s,\pi^*(s)) \in D^\prime \, | \, \pi_{safe} \text{ is unsafe} \}$\;
    Aggregate datasets: $D \gets D \cup D^\prime$\;
	Train policy $\pi_{i+1}$ on $D$\;
 }
\Return{best $\pi_i$ on validation}
\end{algorithm}

While SafeDAGGER has shown promise in various domains, its application to legged robot locomotion control remains largely unexplored. The main reason is that it is difficult to generate expert data for the locomotion problems. However, due to the advancements of the optimal control solvers, nowadays, we can solve NMPC for locomotion efficiently which can act as an expert for training policies. 

\subsection{Guided policy search (GPS)}
GPS uses local optimal control policies around different trajectories to learn control policies \cite{levine2013guided}. The main idea is to first train a policy using BC on the trajectories from the local optimal control policies and then iteratively improve the policy collecting new data from the current policy. The key in the iterations is to guide the policy search to regions of high reward using importance sampling. In this work, our focus is to reduce the number of falls during training, however, it is difficult to include safety considerations in GPS. Nevertheless, similar to GPS, we use MPC as the expert.

\section{Methodology}
\begin{figure*}[htbp]
\let\@latex@error\@gobble
  \begin{multicols}{2}
    \begin{algorithm}[H]
    \caption{LocoDAGGER} \label{alg:mod_dagger}
    \KwIn{Number of algorithm iterations $N$,\newline Data collection rollout episode length $T$,\newline Expert frequency $f_{expert}$ ,\newline Expert influence decay rate $\alpha$,\newline Number of expert queries $N_{query}$,\newline Expert query episode length $T_{query}$}
    \KwOut{Best performing policy $\pi$ on evaluation}
    \vspace{0.3cm}
    \namedblock{Pre-training (Behavior Cloning)}{
        Initialize policy $\pi_0$\;
        $D \gets$ Collect rollout data from expert\;
        Train policy $\pi_0$ on $D$\;
    }
    \SetKwBlock{MainLoop}{\textbf{for} \textit{$i=1$ to $N$} \textbf{do}}{end}
    \tcp{Main Algorithm Loop}
    \MainLoop{
        \namedblock{Policy Rollout}{
            Expert influence $p \gets \alpha^{i-1}$\;
            State history $s \gets \emptyset$\;
            \tcp{Simulation Loop}
            \For{$t=0$ to $T$}{
                \If{$t$ mod $f_{expert} == 0$}{
                    $use\_mpc \gets$ random choice with probability $p$\;
                }
                \eIf{$use\_mpc$ is $True$}{
                    Use expert to control robot\;
                }{
                    Use policy $\pi_{i-1}$ to control robot\;
                }
                $s \gets s \cup s_t$\;
            }
        }
        \namedblock{Data Collection}{
            \For{$j=0$ to $N_{query}$}{
                $s_0 \gets$ random sample from $s$\;
                \tcp{Simulation Loop}
                \For{$t=0$ to $T_{query}$}{
                    Use expert to control robot\;
                    $D \gets D \cup \{\vec{s}_t, \vec{a}_t, \vec{g}_t\}$\;
                }
            }
        }
        \namedblock{Training}{
            Train policy $\pi_i$ on $D$\;
        }
    }
    \Return{Best performing policy $\pi$ on evaluation}
    \end{algorithm}
    
    \columnbreak
    
    \begin{algorithm}[H]
    \caption{LocoSafeDAGGER} \label{alg:mod_SAFEdagger}
    \KwIn{Number of algorithm iterations $N$,\newline Data collection rollout episode length $T$,\newline Safety policy $\pi_{safety}$,\newline Expert rollout episode length if unsafe $T_{block}$,\newline Expert ending rollout episode length $T_{end}$\newline}
    \KwOut{Best performing policy $\pi$ on evaluation}
    \vspace{0.3cm}
    \namedblock{Pre-training (Behavior Cloning)}{
        Initialize policy $\pi_0$\;
        $D \gets$ Collect rollout data from expert\;
        Train policy $\pi_0$ on $D$\;
    }
    \SetKwBlock{MainLoop}{\textbf{for} \textit{$i=1$ to $N$} \textbf{do}}{end}
    \tcp{Main Algorithm Loop}
    \MainLoop{
        \namedblock{Policy Rollout + Data Collection}{
            \tcp{Simulation Loop}
            \For{$t=0$ to $T$}{
                \eIf{$\pi_{safety} \rightarrow$ unsafe}{
                    Use expert to control robot\;
                    $D \gets D \cup \{\vec{s}_t, \vec{a}_t, \vec{g}_t\}$\;
                    Activate blocking for $T_{block}$ episodes\;
                }{
                    \eIf{blocking is complete}{
                        Use policy $\pi_{i-1}$ to control robot\;
                    }{
                        Use expert to control robot\;
                        $D \gets D \cup \{\vec{s}_t, \vec{a}_t, \vec{g}_t\}$\;
                    }                
                }
            }
        }
        \namedblock{Extra Data Collection}{
            \tcp{Simulation Loop}
            \For{$\tau=t$ to $t+T_{end}$}{
                Use expert to control robot\;
                $D \gets D \cup \{\vec{s}_\tau, \vec{a}_\tau, \vec{g}_\tau\}$\;
            }
        }
        \namedblock{Training}{
            Train policy $\pi_i$ on $D$\;
        }
    }
    \Return{Best performing policy $\pi$ on evaluation}
    \end{algorithm}

  \end{multicols}
\end{figure*}

In this section, we introduce new adaptations to the DAGGER and SafeDAGGER algorithms, making them more suited for deployment on legged robots, using MPC as the expert. The algorithms also account for the different operating frequencies of the policy and the MPC. The policy can run at a different frequency than the MPC. For instance, in our example, the policy executed at $1kHz$ and the MPC regenerate trajectories at $20Hz$.

\subsection{LocoDAGGER}
We introduce a modified version of DAGGER, namely LocoDAGGER, that adapts the original algorithm to be used with an MPC controller (see algorithm \ref{alg:mod_dagger}). To improve the stability and learning efficiency of the policy during early iterations, we introduce a pre-training stage before the main algorithm loop. There, the dataset is preloaded with data from a few expert rollouts, and the policy is pre-trained on this dataset. The main algorithm loop consists of three stages: policy rollout, data collection, and training. In line with the original DAGGER algorithm, we first calculate the expert influence probability $p$, which decays exponentially with progressing algorithm iterations with the decay rate $\alpha$. The early stages of training are dominated by the expert rollouts, while transitioning slowly to the rollouts from the learned policy. At each expert frequency $f_{expert}$ step during the policy rollout, we select randomly, with the calculated expert influence probability $p$, whether the expert or the policy controls the robot. The robot states during the rollout are recorded. Then, during the data collection stage, the expert is deployed from randomly sampled points along the recorded robot state trajectory, and its actions, along with the states and goals, are added to the dataset. This process is repeated for a given number of expert queries $N_{query}$. The policy is then retrained on the entire dataset, and the main algorithm loop repeats. For each iteration, the policy is trained with significantly fewer epochs than BC to prevent overfitting. Overall, this approach reduces the number of expert queries compared to the original version, as querying every single state in the trajectory is too computationally expensive.

\subsection{LocoSafeDAGGER}
The algorithm in the previous section is indifferent to the number of failures during training. However, it is always desired to learn policies directly in the real world, which requires a minimum number of falls during training. To address this, we propose LocoSafeDAGGER (see Algorithm \ref{alg:mod_SAFEdagger}). Similar to LocoDAGGER, the algorithm starts with a pre-training stage to increase stability, where the dataset is preloaded with expert data, and the policy is pre-trained on this dataset. The main algorithm loop also consists of three stages: a combined policy rollout and data collection stage, an extra data collection stage, and a training stage. During the policy rollout in the first stage, a safety policy $\pi_{safety}$ monitors various aspects of the robot, for example, its base position and angles, and joint configurations. If the safety condition is violated, the expert takes over for a fixed duration $T_{block}$ (blocking). This blocking step is necessary to allow the expert time to bring the robot back into a stable moving state. Control will be handed back to the policy once the robot state is deemed safe again or the blocking duration is completed. The dataset will only include information from periods when the expert is in control of the system. Specifically, we will record the system states, goals, and actions taken during these expert-controlled periods. Any data generated during periods when the policy is in control will be excluded from the dataset. To address scenarios where the robot remains safe but exhibits suboptimal behaviors (e.g., standing still without taking steps), we incorporate an additional data collection phase after the policy rollout. In this phase, the expert is deployed from the final recorded robot state for a fixed duration $T_{end}$. Finally, the policy is trained on the entire dataset and the main algorithm loop repeats. 

\section{Implementation}
The experiments in this work were conducted using a simulation model of the Solo12 quadruped \cite{Grimminger.2020} in a PyBullet environment \cite{ErwinCoumans.20162021}. The data collection, simulation, and training were done on a workstation equipped with a 13th Gen Intel(R) Core(TM) i7-13700K CPU, 64GB RAM, and an NVIDIA GeForce RTX 4090 GPU. All experiments utilized the trot gait, with a desired CoM velocity ranging from $0.0\frac{m}{s}$ to $0.3\frac{m}{s}$ in the positive $x$ direction. This limited gait choice and desired velocity range is chosen as the main objective is to compare different algorithms in a standard setting. Since the data collection and evaluation processes involves sampling from a distribution and has a probabilistic nature, a total of 3 runs per learning technique were performed to ensure that the results are not merely due to chance. 

\subsection{Policy Network}
The policy is parameterized with a fully connected MLP network with 3 hidden layers, each comprising 512 nodes. $1D$ batch normalization is performed between each layer, followed by a ReLU activation function. Training of the network utilizes a $L1$ loss function on the network output, and all networks are trained with a learning rate of 0.002 and a batch size of 256, with different epochs depending on the learning technique. Initialization of all networks follows the Kaiming initialization method \cite{He.262015}. The network's input consists of the robot states and desired goal, both separately normalized. To simplify the problem, it is assumed that the robot has full observation of all states. The robot states include:
\begin{itemize}
    \item Base orientation and velocity
    \item Joint positions and velocities
    \item Relative $x, y$ positions of the end effectors to the robot CoM
\end{itemize}
The desired goal for each simulation step encompasses:
\begin{itemize}
    \item Desired gait phase
    \item Desired CoM $x, y$ and robot base yaw velocity
\end{itemize}
The network outputs the desired joint position target as the action, which has been shown to outperform other possibilities, such as direct torque output \cite{Khadiv.2023}. These desired positions are then passed to a PD controller with low impedance to compute the joint torques used to control the robot. 

\subsection{Robot Initial Condition Generation}
To increase the complexity of the problem and to collect a more diverse dataset that includes recovery maneuvers, the robot's initial configuration is randomized with contact-consistent perturbations \cite{Khadiv.2023} along the nominal trajectory for all rollouts. 

\subsection{The Expert}
The expert used to control the quadruped in this work is the NMPC in \cite{Meduri.19012022}. i.e., BiConMP. It splits the problem into kinematic and dynamic optimal control problems. Given a contact plan for the robot's end-effectors, the dynamic and kinematic optimization problem are solved sequentially. the plans are updated at 20 Hz given the current state of the robot in a closed-loop MPC fashion, and the trajectories are passed to an unconstrained inverse dynamics module to generate joint torques at $1kHz$.

\subsection{Iterative Algorithm Parameters}

In the experiments, the LocoDAGGER was pre-trained with $112k$ data points. These data points were collected from $100 \times 1.5s$ rollouts from 10 different velocities. 10 iterations of the algorithm were performed, with the policies being trained for 15 epochs per iteration. A total of 10 policy rollouts are performed per iteration, each spanning $5s$. Then, an average of 5 samples are taken per trajectory for the NMPC rollout, which lasted $1s$ per sample. The NMPC influence decay rates of $0.5$ and $0.8$ were investigated. 

Similar to the LocoDAGGER, the same $112k$ data points were also used to pre-train the policy during the experiments of LocoSafeDAGGER. The safety policy monitors specific robot states, ensuring they remain within boundaries where the robot is considered safe for policy control, as detailed in table \ref{tbl:safety_policy}. The algorithm was also run for 10 iterations. For each iteration, a total of $40 \times 10s$ policy rollouts were performed. The NMPC is set to take over for at least $2s$ if the robot is unsafe, and an extra data collection step at the end of each policy rollout is done with a $1s$ NMPC rollout. A total of 15 training epochs were done per iteration.

\begin{table}[htbp]
\centering
\begin{tabular}{|cc|c|}
\hline
\multicolumn{2}{|l|}{}                                                 & \multicolumn{1}{c|}{Bounds} \\ \hline
\multicolumn{2}{|c|}{CoM Height}                                       & $[0.15\,m, 1.0\,m]$                            \\ \hline
\multicolumn{2}{|c|}{Body Roll and Pitch}                              & $[-25^\circ, 25^\circ]$                            \\ \hline
\multicolumn{1}{|c|}{\multirow{2}{*}{Hip Abduction/Adduction}} & Left  & $[-30^\circ, 60^\circ]$                            \\ \cline{2-3} 
\multicolumn{1}{|c|}{}                                         & Right & $[-60^\circ, 30^\circ]$                            \\ \hline
\multicolumn{1}{|c|}{\multirow{2}{*}{Hip Flexion/Extension}}   & Front & $[0^\circ, 90^\circ]$                            \\ \cline{2-3} 
\multicolumn{1}{|c|}{}                                         & Back  & $[-90^\circ, 0^\circ]$                            \\ \hline
\multicolumn{1}{|c|}{\multirow{2}{*}{Knee Flexion/Extension}}  & Front & $[-160^\circ, 0^\circ]$                            \\ \cline{2-3} 
\multicolumn{1}{|c|}{}                                         & Back  & $[0^\circ, 160^\circ]$                            \\ \hline
\end{tabular}
\caption{Safety Policy with safe robot state bounds} \label{tbl:safety_policy}
\end{table}

\subsection{Evaluation Procedure}
To evaluate the safety aspect of the iterative algorithms during data collection, we recorded the failure rate of the robot during training. For LocoDAGGER, a data collection rollout is deemed failed if the NMPC diverges while in control, or if the robot falls over while the policy is in control. For LocoSafeDAGGER, a data collection rollout is considered a failure if the NMPC diverges.

To better evaluate the performance of the iterative algorithms, BC was used as the benchmark. To facilitate comparison with iterative algorithms, datasets of varying sizes were collected. In this work, an average maximum of $736k$ samples were collected from 20 different desired velocities, with a total of $400 \times 3s$ rollouts. The policy networks were then trained on the collected datasets for 150 epochs.

For each trained policy from BC and for the policy from each iteration of the iterative algorithms, an evaluation of its robustness (against perturbed initial conditions) and performance (desired velocity tracking) was conducted.the experiments, a total of $100 \times 5s$ evaluation rollouts were performed for each policy, which includes 10 equally spaced velocities spanning the entire desired velocity range. A rollout is considered successful if the robot does not fall over and maintains a constant stepping motion for at least $2/3$ of the total episode time. For the tracking performance, the mean squared error (MSE) between the actual and desired base $x, y$, and yaw velocity ($v_x, v_y, \omega$) was used.

Finally, the best-performing policy for each learning technique was tested for push recovery robustness. This test involved finding the maximum impulse the policy could withstand from all directions, measured at $45^\circ$ intervals. A policy was considered robust against a certain impulse if it achieved a rollout success rate of $\geq80\%$.

\section{Results and Discussions}
This section presents and analyzes the outcomes of the experiments with different learning techniques for quadrupedal locomotion control. The evaluation encompasses three key dimensions, each addressed in a separate subsection:

\begin{itemize}
    \item Safety During Data Collection: An assessment of the safety (avoiding failures) during the data collection process for the iterative algorithms.
    \item Policy Performance: An examination of the trained policies' robustness to perturbations and their ability to accurately track desired velocities.
    \item Push Recovery Capabilities: For the best-performing policies from each technique, an investigation of their ability to maintain balance under external disturbances.
\end{itemize}


\subsection{Safety during Data Collection}
\begin{figure}[htbp]
  \centering
  \includegraphics[width=0.4\textwidth]{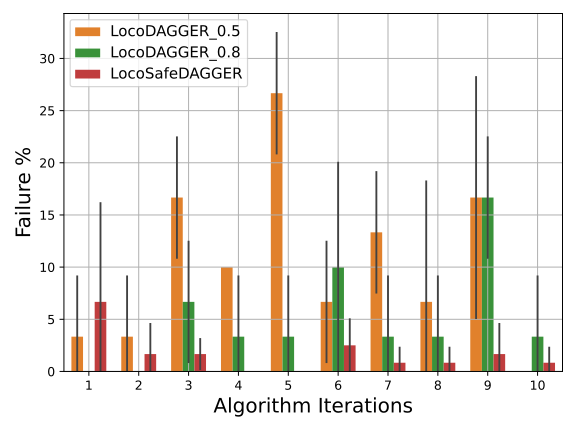}
  \caption{Rollout failure rates during data collection. LocoSafeDAGGER maintains a consistently low failure rate ($<5\%$), while LocoDAGGER shows higher failure rates ($>5\%$) in later iterations as expert influence decreases.}
  \label{fig:data_collection_failure_rate}
\end{figure}

\begin{figure}[htbp]
  \centering
  \includegraphics[width=0.4\textwidth]{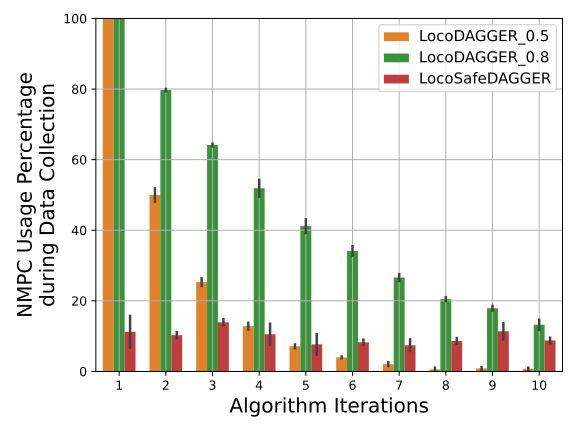}
  \caption{NMPC usage during data collection. LocoSafeDAGGER maintains a consistent expert query rate of around 10\%, while LocoDAGGER's rate decreases monotonically from 100\%. LocoSafeDAGGER achieves lower computational demands without compromising data collection safety.}
  \label{fig:nmpc_influence}
\end{figure}

The percentage of failed rollouts during data collection is depicted in Figure \ref{fig:data_collection_failure_rate}. It can be observed that LocoSafeDAGGER maintains a very low failure rate during data collection, with the failure rate exceeding 5\% only in the first iteration. In contrast, while the strong expert influence in the LocoDAGGER algorithm results in stable data collection rollouts during the early iterations, failure rates significantly increase in later iterations. This trend is particularly pronounced for the LocoDAGGER algorithm with a 0.5 influence decay rate, where the rollout failure rate consistently exceeds 5\% after the second iteration. These results demonstrate that the active safety check of LocoSafeDAGGER enhances overall rollout safety during data collection. Notably, the rollouts for LocoSafeDAGGER are twice as long as those for LocoDAGGER in the experiments, further highlighting the safety capabilities of the LocoSafeDAGGER algorithm. This increased safety did not come at the cost of additional expert queries, as observed in figure \ref{fig:nmpc_influence}. The figure shows that the NMPC (Nonlinear Model Predictive Control) influence in the LocoDAGGER algorithm exhibits a monotonic decay from 100\%, whereas the NMPC usage remains relatively constant at around 10\% across all iterations in LocoSafeDAGGER. LocoSafeDAGGER's efficient use of the expert, in contrast to LocoDAGGER's predetermined expert influence rate, results in significantly fewer expert queries, thereby increasing computational efficiency during data collection.

\subsection{Policy Robustness and Tracking Performance}
\begin{figure}[htbp]
  \centering
  \includegraphics[width=0.4\textwidth]{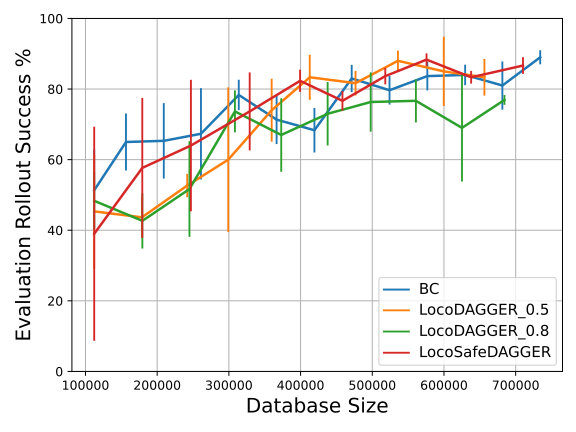}
  \caption{Policy robustness. All approaches except LocoDAGGER with high expert influence achieve a similar evaluation rollout success rate of about 90\%, demonstrating that iterative algorithms can match the policy robustness of BC.}
  \label{fig:robustness}
\end{figure}

The robustness of trained policies against initial condition perturbations is illustrated in figure \ref{fig:robustness}. For better comparison with BC, the iterations of the iterative algorithms are mapped to their respective database sizes. The figure demonstrates that policy robustness increases with database size for all learning techniques. All approaches, except LocoDAGGER with 0.8 NMPC influence decay rate, show robustness plateauing just below 90\%.

\begin{figure*}
  \centering
  \includegraphics[width=\textwidth]{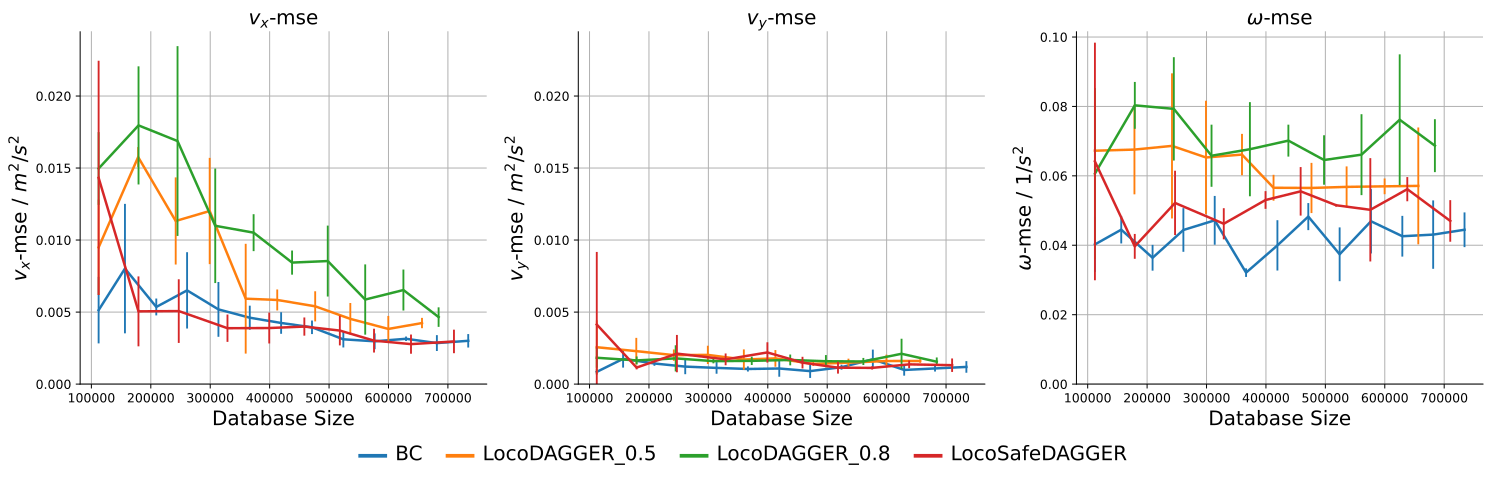}
  \caption{Velocity tracking MSE. BC demonstrates the best overall velocity tracking performance, with LocoSafeDAGGER almost matching BC's tracking performance. LocoDAGGER with high expert interference leads to overly conservative data collection rollouts, resulting in lower tracking performance.}
  \label{fig:vel_tracking_mse}
\end{figure*}

Figure \ref{fig:vel_tracking_mse} depicts the velocity tracking mean squared error (MSE) for increasing database sizes. LocoDAGGER with a decay rate of 0.8 performs the worst in both $v_x$ and $\omega$ tracking. For $v_x$ tracking, BC and LocoSafeDAGGER converge to the same error value, while LocoDAGGER with 0.5 NMPC influence decay rate converges to a slightly higher value and exhibits higher errors in earlier iterations. BC demonstrates the best performance in $\omega$ tracking, followed by LocoSafeDAGGER and then LocoDAGGER. No clear best performer emerges for $v_y$ tracking.

Based on these results, it can be deduced that using high expert influence for LocoDAGGER does not produce robust policies with good tracking behavior. This may be attributed to overly conservative data collection policy rollouts with high expert interference, which inhibit real exploration and nullify the core advantage of online learning methods. LocoSafeDAGGER, in particular, demonstrated better sample efficiency than LocoDAGGER, achieving the same database size with four times the number of policy rollouts. The results also demonstrate that iterative algorithms can produce outcomes comparable to classical BC, showing no significant disadvantage in terms of policy robustness and tracking performance. 

\subsection{Push Recovery Robustness}

\begin{figure}[htbp]
  \centering
  \includegraphics[width=0.45\textwidth]{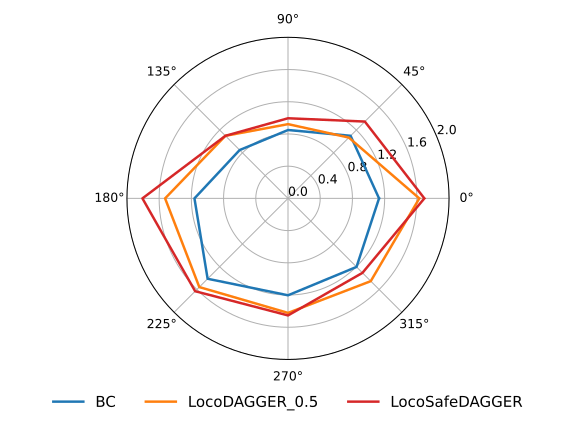}
  \caption{Average maximum impulse $[Ns]$ tolerated from all directions with $\geq80\%$ successful recovery rate. Iterative algorithms' data collection strategies produce datasets more representative of the trained policy's encountered distribution during rollout, enhancing policy robustness against external disturbances.}
  \label{fig:push_recovery}
\end{figure}

To test the robustness of the resulting policies from each method in terms of push recovery, we picked the best-performing policy from each method. Figure \ref{fig:push_recovery} illustrates the mean maximum impulse policies from each approach can withstand while maintaining at least an 80\% recovery success rate. The results indicate that BC exhibits poorer push recovery performance compared to the two iterative algorithms.

BC's inferior performance can be attributed to its data collection method. It only collects data from expert rollouts. Since the expert is less prone to mistakes than an untrained policy, the resulting dataset does not represent the data distribution encountered by the trained policy during execution. rmore, LocoSafeDAGGER demonstrates slightly better performance than LocoDAGGER for most impulse angles. This difference can be attributed to the distinct data collection strategies employed by the two algorithms. LocoSafeDAGGER specifically collects data in dangerous states, in contrast to the random states used in LocoDAGGER. Consequently, LocoSafeDAGGER is more likely to gather data that is more relevant to critical scenarios, resulting in a more robust policy.

\section{Conclusion and Future Works}
This study presents a novel algorithm for learning policies from MPC experts that takes the safety of the robot during data collection into account. We performed an extensive comparative evaluation of different algorithms, specifically focusing on safety during data collection and the robustness and performance of the trained policies. The results showed that the LocoSafeDAGGER algorithm significantly enhances rollout safety during data collection, maintaining low failure rates and demonstrating high sampling efficiency without increasing the number of expert queries required. In terms of policy robustness and tracking performance, it was observed that all learning techniques, except for LocoDAGGER with a high expert influence decay rate, showed robustness improvements with increasing database size. Notably, LocoSafeDAGGER performed comparably to BC in velocity tracking and outperformed LocoDAGGER in robustness against initial condition perturbations. The push recovery tests revealed that iterative algorithms, including LocoSafeDAGGER, exhibited superior performance compared to BC. This advantage is likely due to the iterative algorithms' ability to explore and collect data more representative of the states encountered during actual rollouts, thus enhancing the dataset's relevance and the trained policy's effectiveness.

This work opens several promising avenues for future research in legged robot control. A key area for improvement lies in enhancing the safety mechanisms of LocoSafeDAGGER. Implementing a neural network that processes current and historical sensor data could potentially predict future failures can complement the LocoSafeDAGGER in allowing MPC to take over earlier. The safety policy could also be gradually relaxed as the algorithm progresses, allowing for increased policy exploration and enhancing the diversity of the collected dataset. Future work will also extend the experimental scope to encompass a wider range of gaits, target velocities, and more challenging terrains. Introducing sensor noise and external forces during data collection would provide a more comprehensive evaluation of the algorithms' robustness and adaptability. A crucial next step would be transferring these results to real robots, allowing for direct comparison of the performance, robustness, and adaptability of policies trained using iterative algorithms against those developed through behavior cloning and reinforcement learning.


\bibliography{literature} 

\begin{thebibliography}{10}

\bibitem{Meduri.19012022}
A.~Meduri, P.~Shah, J.~Viereck, M.~Khadiv, I.~Havoutis, and L.~Righetti, ``Biconmp: A nonlinear model predictive control framework for whole body motion planning.''

\bibitem{mastalli2023agile}
C.~Mastalli, W.~Merkt, G.~Xin, J.~Shim, M.~Mistry, I.~Havoutis, and S.~Vijayakumar, ``Agile maneuvers in legged robots: a predictive control approach,'' {\em IEEE Transactions on Robotics}, 2023.

\bibitem{grandia2023perceptive}
R.~Grandia, F.~Jenelten, S.~Yang, F.~Farshidian, and M.~Hutter, ``Perceptive locomotion through nonlinear model predictive control,'' {\em IEEE Transactions on Robotics}, 2023.

\bibitem{tassa2011stochastic}
Y.~Tassa and E.~Todorov, ``Stochastic complementarity for local control of discontinuous dynamics,'' in {\em Proceedings of Robotics: Science and Systems}, 2011.

\bibitem{drnach2021robust}
L.~Drnach and Y.~Zhao, ``Robust trajectory optimization over uncertain terrain with stochastic complementarity,'' {\em IEEE Robotics and Automation Letters}, vol.~6, no.~2, pp.~1168--1175, 2021.

\bibitem{hammoud2021impedance}
B.~Hammoud, M.~Khadiv, and L.~Righetti, ``Impedance optimization for uncertain contact interactions through risk sensitive optimal control,'' {\em IEEE Robotics and Automation Letters}, vol.~6, no.~3, pp.~4766--4773, 2021.

\bibitem{ha2024learning}
S.~Ha, J.~Lee, M.~van~de Panne, Z.~Xie, W.~Yu, and M.~Khadiv, ``Learning-based legged locomotion; state of the art and future perspectives,'' {\em arXiv preprint arXiv:2406.01152}, 2024.

\bibitem{carius2020mpc}
J.~Carius, F.~Farshidian, and M.~Hutter, ``Mpc-net: A first principles guided policy search,'' {\em IEEE Robotics and Automation Letters}, vol.~5, no.~2, pp.~2897--2904, 2020.

\bibitem{viereck2022valuenetqp}
J.~Viereck, A.~Meduri, and L.~Righetti, ``Valuenetqp: Learned one-step optimal control for legged locomotion,'' in {\em Learning for Dynamics and Control Conference}, pp.~931--942, PMLR, 2022.

\bibitem{Bogdanovic.7142021}
M.~Bogdanovic, M.~Khadiv, and L.~Righetti, ``Model-free reinforcement learning for robust locomotion using demonstrations from trajectory optimization.''

\bibitem{Khadiv.2023}
M.~Khadiv, A.~Meduri, H.~Zhu, L.~Righetti, and B.~Sch{\"o}lkopf, ``Learning locomotion skills from mpc in sensor space,'' {\em Proceedings of The 5th Annual Learning for Dynamics and Control Conference}, vol.~211, pp.~1218--1230, 2023.

\bibitem{jenelten2024dtc}
F.~Jenelten, J.~He, F.~Farshidian, and M.~Hutter, ``Dtc: Deep tracking control,'' {\em Science Robotics}, vol.~9, no.~86, p.~eadh5401, 2024.

\bibitem{Ross.2010}
S.~Ross and D.~Bagnell, ``Efficient reductions for imitation learning,'' in {\em Proceedings of the Thirteenth International Conference on Artificial Intelligence and Statistics} (Y.~W. Teh and M.~Titterington, eds.), vol.~9 of {\em Proceedings of Machine Learning Research}, (Chia Laguna Resort, Sardinia, Italy), pp.~661--668, PMLR, 2010.

\bibitem{pomerleau1988alvinn}
D.~A. Pomerleau, ``Alvinn: An autonomous land vehicle in a neural network,'' {\em Advances in neural information processing systems}, vol.~1, 1988.

\bibitem{Hussein.2018}
A.~Hussein, M.~M. Gaber, E.~Elyan, and C.~Jayne, ``Imitation learning: A survey of learning methods,'' {\em ACM Computing Surveys}, vol.~50, no.~2, pp.~1--35, 2018.

\bibitem{youm2023imitating}
D.~Youm, H.~Jung, H.~Kim, J.~Hwangbo, H.-W. Park, and S.~Ha, ``Imitating and finetuning model predictive control for robust and symmetric quadrupedal locomotion,'' {\em IEEE Robotics and Automation Letters}, 2023.

\bibitem{Ross.1122010}
S.~Ross, G.~J. Gordon, and J.~A. Bagnell, ``A reduction of imitation learning and structured prediction to no-regret online learning.''

\bibitem{Zhang.5202016}
J.~Zhang and K.~Cho, ``Query-efficient imitation learning for end-to-end autonomous driving.''

\bibitem{levine2013guided}
S.~Levine and V.~Koltun, ``Guided policy search,'' in {\em International conference on machine learning}, pp.~1--9, PMLR, 2013.

\bibitem{Grimminger.2020}
F.~Grimminger, A.~Meduri, M.~Khadiv, J.~Viereck, M.~W{\"u}thrich, M.~Naveau, V.~Berenz, S.~Heim, F.~Widmaier, T.~Flayols, J.~Fiene, A.~Badri-Spr{\"o}witz, and L.~Righetti, ``An open torque-controlled modular robot architecture for legged locomotion research,'' {\em IEEE Robotics and Automation Letters}, vol.~5, no.~2, pp.~3650--3657, 2020.

\bibitem{ErwinCoumans.20162021}
{Erwin Coumans} and {Yunfei Bai}, ``Pybullet, a python module for physics simulation for games, robotics and machine learning,'' 2016--2021.

\bibitem{He.262015}
K.~He, X.~Zhang, S.~Ren, and J.~Sun, ``Delving deep into rectifiers: Surpassing human-level performance on imagenet classification.''

\end{thebibliography}
\bibliographystyle{ieeetr}

\end{document}